\documentclass[10pt]{article}
\usepackage[preprint]{tmlr}
\usepackage{hyperref}
\usepackage{url}
\usepackage{graphicx}
\usepackage{booktabs}
\usepackage{amsmath,amssymb}
\usepackage{array}

\title{On-Device Deep Research at 4B: Exposure Bounds\\ Faithfulness, Retrieval Bounds Coverage}

\author{\name Vinay Kumar Chaganti \email cvk.atreya@gmail.com \\
      \addr Independent Researcher}

\newcommand{\availabilitytext}{The paper list that defines the corpus, deterministic scripts that rebuild it from public sources, and the per-run records and per-claim grades that regenerate every table and figure are released. Source article text is not redistributed; it is rebuilt locally. The code and data will be released upon publication.}

\begin{document}
\maketitle

\begin{abstract}
On-device research agents search a corpus, read sources, and write a cited brief on a personal laptop.
Whether their citations are faithful, and at what cost, is unmeasured for a deployable small model. This
study fixes one 4B generator on a 24 GB laptop and asks what makes its citations faithful. It separates
two quantities usually reported as one number. Cited claim faithfulness asks whether the cited source
supports the claim. Trustworthy coverage asks whether the agent also cites the right sources. The study
crosses how much of each source the generator sees, 400 against 1500 characters, with the quality of the
sources supplied, gold papers against retrieved papers. Two levers fall out, and they act on different
outcomes. Exposure sets faithfulness. More of each source lifts faithfulness from 0.45 to 0.58 on retrieved
sources and from 0.37 to 0.58 on gold sources, and the two settings converge, so faithfulness is bound
by exposure, not by whether the source is correct. The exposure lift is robust to a second, independent
judge; the exact convergence is tight under the primary judge and only approximate under the second.
Retrieval sets coverage. Trustworthy coverage stays
near 0.22 on retrieved sources at any exposure, because recall is held near 0.40, so exposure cannot fix
which sources are cited. The extra exposure costs about 235 output tokens. The practical recipe is to
raise per source exposure first, cheaply, and then treat retrieval recall as the only remaining lever.
\end{abstract}

\section{Introduction}
Deep research agents search sources, read them, and write a cited report. They are
evaluated mainly as commercial products, compared on answer accuracy at frontier
scale~\citep{deepresearchbench,researcherbench,browsecomp}. Two questions decide whether such an agent is
usable on a personal device. First, are its citations faithful, meaning does each cited source actually
support the claim attached to it. Second, what does it cost on hardware a practitioner owns. The field
increasingly names the first gap. Surveys report a mismatch between answer accuracy scores and the real
objectives of research agents~\citep{drroadmap,howfaruseful}. Studies of attribution find that citations
often resolve yet fail to support their claims~\citep{citednotverified}. Uncontrolled web evaluation is
shown to inflate scores by retrieving benchmark answers at inference
time~\citep{searchtimecontamination}. Cost is named as underreported across the board~\citep{agenticragsurvey,efficientbench}.

This study fixes one 4B generator, small enough to run on a single 24 GB laptop, and asks a plain
question. What makes its citations faithful, and what does the fix cost. The model is held fixed on
purpose. The question is not which size to pick. It is what design makes a fixed, deployable size
faithful.

The central move is to stop reporting citation quality as one number. An agent can support the sources it
cites yet cite the wrong ones. It can also cite the right sources yet fail to support them. The study
measures these separately. \emph{Cited claim faithfulness} is the share of cited claims whose cited
source supports them. \emph{Trustworthy coverage} is faithfulness weighted by how much of the gold
evidence the agent actually cites~\citep{faithbyconstruction}. Keeping them apart is what makes the result
legible.

The result is a clean split into two levers that act on different outcomes. The first lever is
\emph{exposure}, how much of each source the generator sees. Raising exposure from 400 to 1500 characters
per source lifts faithfulness from 0.45 to 0.58 on retrieved sources, and from 0.37 to 0.58 on gold
sources. The two settings converge at high exposure. Faithfulness is therefore bound by exposure, not by
whether the cited source is the correct one. The second lever is \emph{retrieval}. Trustworthy coverage
stays near 0.22 on retrieved sources at either exposure, because citation recall is held near 0.40 by
the retrieval stage. Exposure does not touch it. The extra exposure is nearly free, about 235 output
tokens. The deployable recipe follows directly. Raise per source exposure first, because it is cheap and
it maxes out faithfulness. Then the only lever left for trustworthy coverage is retrieval recall.

\section{Related Work}
\subsection{Deep research agents and their evaluation}
Benchmarks for research agents center on answer accuracy and holistic report quality at frontier
scale~\citep{deepresearchbench,researcherbench,browsecomp,deepscholarbench}. A parallel line builds
controlled substrates: a frozen corpus and a fixed retriever, so scores are reproducible and comparable~\citep{deepresearchgym,browsecompplus}. This study adopts that substrate discipline and adds the
part those benchmarks omit, a fixed generator below 8B scored on citation faithfulness and cost on a
laptop.

\subsection{Citation faithfulness and attribution}
Claim-level citation precision and recall come from ALCE~\citep{alce}, with a long form judge from
LongCite~\citep{longcite}. Attribution work shows citations that resolve but do not support their
claims~\citep{citednotverified}, motivating checking each claim again against its cited source. Small encoder
verifiers such as MiniCheck~\citep{minicheck} and HHEM~\citep{hhem} match large judges on grounding at a
fraction of the cost, and are used here only to score, not to train. Trustworthy coverage, faithfulness
weighted by recall, is used as a reporting lens following recent work that couples the two~\citep{faithbyconstruction}.

\subsection{How much source to show, and how long to write}
Whether showing more of a source helps grounding is contested. WebCiteS finds that extending from curated
snippets to full content lowers citation F1, because more raw text dilutes attribution~\citep{webcites}.
Matched-evidence analysis on several local small models finds that the same evidence embedded in long
input is recovered less well than when supplied compactly~\citep{evidutil}. On the output side, longer
responses show lower factual precision, attributed to the model exhausting its reliable
knowledge~\citep{lengthfactuality}, and grounded long outputs lose faithfulness toward their
end~\citep{hallucinatelast}. These results predict that more exposure may plateau or invert for a small
model. This study measures the actual shape for a fixed 4B and finds a rise, not an inversion, over the
range tested, which is why the range and the model class are stated plainly.

\subsection{Small models and context use}
Models below 7B are reported to underuse retrieved context: with an oracle passage they still miss the
answer often, and added context can overwrite what the model knew~\citep{contextutil}. That work varies
model size with a fixed single passage and scores extractive answer accuracy. It is used here as a foil.
The exposure lever varies how much of each source is shown and scores citation faithfulness, a different
axis, and finds that the 4B does convert more per source exposure into more faithful grounding.

\subsection{Verification in the loop and long form generation}
Two habits keep citations honest. One scores citations after the brief is written~\citep{alce}. The other
asks the generator to critique and revise its own work~\citep{ever,rarr,vericite}. Both lean on the
generator, which is the weak point below 8B: small models correct themselves only with a strong external
checker~\citep{scorecorrect}, cannot reliably correct their own reasoning~\citep{selfcorrectillusion,selfcorrectsurvey}, and are poorly
calibrated~\citep{calibration}. A short output panel here tests these repair habits directly. For grounded
long form generation, Deep-Reporter uses recurrent context management~\citep{deepreporter}, but it
compresses the model's own output draft rather than varying per source input exposure, uses 8B to 70B
backbones, and relies on custom post training. The lever studied here is different and cheaper: change how
much of each source a fixed, untrained 4B reads.

\section{Experimental Design}
\begin{figure}[t]
\centering
\includegraphics[width=0.92\textwidth]{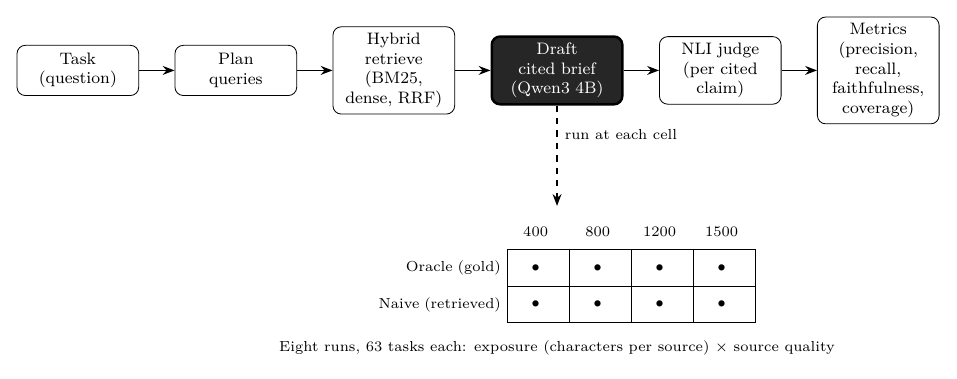}
\caption{The agent and the design. The fixed pipeline runs task, plan, hybrid retrieve, draft, and judge.
The draft step, in the dark box, is where the two controlled factors apply: source quality, oracle gold
papers against naive retrieved papers, and exposure, 400 to 1500 characters shown per source. The draft
step is run at every cell of the grid below, four exposures by two source qualities, giving eight runs of
63 tasks each.}
\label{fig:pipeline}
\end{figure}
\subsection{Corpus, tasks, and gold}
The substrate is built from real papers, so the gold is not invented. The study starts from 63 recent
arXiv papers in machine learning and language technology, dated 2025. Each source paper defines one task.
Its title and abstract become the research question, and the agent must write a cited brief that answers
it. The gold evidence for a task is the set of papers that source paper actually cited, taken from its own
reference list. Gold is curated by the authors of real papers, not by the study. It averages 17.7 papers
per task, ranging from 2 to 35.

The corpus is the union of the 63 source papers and every paper they cite, 1062 documents in all, frozen
as one snapshot with a content hash. Where a cited paper is on arXiv and fetchable, its full body is
stored. In total 656 documents carry full text, and the rest carry a title, an abstract, and a snippet.
Every gold citation resolves to a corpus document by arXiv identifier or by normalized title, so the gold
set is a subset of the corpus by construction. Coverage of the gold is complete. Recall can be lost only
in retrieval, or in what the agent chooses to cite, never because the evidence is absent from the corpus.
That property is what lets the study blame a recall loss on a stage rather than on the data.

\subsection{Models and retriever}
All runs use one generator, Qwen3-4B-Instruct-2507 at 8 bit, served by Ollama on an Apple M4 Pro with
24 GB of memory, at temperature 0.2, a context window of 8192, and a fixed seed. The retriever is hybrid,
BM25 with a nomic dense encoder, fused by reciprocal rank fusion. The scoring judge is a DeBERTa-v3 NLI
cross encoder with a SummaC windowed maximum. It comes from a different model family than the verifier
used inside any repair loop, so the tool that edits a brief never grades it. Table~\ref{tab:config} lists
these settings.

\begin{table}[t]
\centering
\caption{Run configuration. The generator, retriever, judge, and decoding are fixed across all eight runs.
Two factors vary, giving the eight runs of Fig.~\ref{fig:levers}.}
\label{tab:config}
\renewcommand{\arraystretch}{1.15}
\begin{tabular}{@{}ll@{}}
\toprule
Setting & Value \\
\midrule
Generator        & Qwen3-4B-Instruct-2507, 8 bit, Ollama \\
Hardware         & Apple M4 Pro, 24 GB \\
Retriever        & BM25 and nomic dense, RRF fusion \\
Judge            & DeBERTa-v3 NLI, SummaC windowed max \\
Decoding         & temperature 0.2, context 8192, fixed seed \\
Output budget    & 2000 tokens (long) \\
Tasks per run    & 63 \\
\midrule
Factor: source  & oracle (gold) \emph{or} naive (retrieved) \\
Factor: exposure & 400, 800, 1200, or 1500 characters per source \\
\bottomrule
\end{tabular}
\end{table}

\subsection{Pipeline and arms}
The pipeline is shown in Fig.~\ref{fig:pipeline}. The agent plans queries from the task, retrieves, and
drafts a brief in which each sentence cites the sources it draws on by label. Two arms set the quality of
the sources the generator may cite. The
\emph{naive} arm cites from what retrieval surfaced. The \emph{oracle} arm is handed the gold papers.
Comparing the two isolates the effect of source quality while everything else is held fixed. A separate
short output panel adds arms that repair the brief after generation, described in
Section~\ref{sec:repair}: a check after the fact, a gate that drops unsupported claims, a revise step that
rewrites weak claims, and a reattribute step that reattaches their citations.

\subsection{The exposure lever and the design grid}
Exposure is how many characters of each source the generator sees while drafting, a plain truncation
budget on the same sources, held apart from retrieval. The main experiment crosses exposure, at 400, 800,
1200, and 1500 characters per source, with source quality, oracle against naive, at long output, a draft
budget of 2000 tokens. The resulting grid of eight runs asks whether the exposure effect survives on
retrieved sources, and whether it depends on having the right sources.

\subsection{Metrics}
Let $C$ be the set of cited sources, $G$ the gold set, and $R$ the set retrieval surfaced.
Table~\ref{tab:metrics} lists every metric and its formula. Three points set them apart. First,
\emph{faithfulness} is relative to the cited source. A claim counts as supported when the judge scores an entailment
probability of at least 0.5 between the claim and its \emph{cited} source, so faithfulness asks whether
the agent supports what it chose to cite, not whether it cited the gold. This is why a naive arm can post
high faithfulness while citing the wrong sources. Second, citation recall factors cleanly into
the retrieval and generator terms through the identity in the table, which localizes where recall is lost.
Third, \emph{trustworthy coverage} multiplies faithfulness by recall, so it rewards only claims that are
both supported and correctly attributed, and it is the one number to compare across arms.

\begin{table}[t]
\centering
\caption{Metrics. Faithfulness is relative to the cited source. Trustworthy coverage is the headline number for comparing arms.}
\label{tab:metrics}
\renewcommand{\arraystretch}{1.3}
\begin{tabular}{@{}ll@{}}
\toprule
Metric & Formula \\
\midrule
Citation precision   & $|C \cap G| / |C|$ \\
Citation recall      & $|C \cap G| / |G|$ \\
Retrieval recall     & $|R \cap G| / |G|$ \\
Utilization          & $|C \cap G| / |R \cap G|$ \\
Recall identity      & $\text{cit. recall} = \text{ret. recall} \times \text{util.}$ \\
Faithfulness         & $n_{\text{supported}} / n_{\text{claims}}$ \\
Trustworthy coverage & $\text{faithfulness} \times \text{citation recall}$ \\
Uncited claim rate   & uncited sentences $/$ all sentences \\
\bottomrule
\end{tabular}
\end{table}

What the metrics cover is source support and attribution: whether a cited source entails its claim, and
whether the cited set matches the gold set. What they do not cover is human judged correctness, since
support is scored by an NLI model, not a person, and fluency, coherence, or answer accuracy, which are
not measured here. Every run stores raw per-claim grades with judge identity and provenance.

\subsection{Statistical testing}
The eight runs share the same 63 tasks, so every comparison is paired by task. Each comparison uses the
Wilcoxon signed-rank test, two sided, on the per-task metric, which suits a bounded rate that is not
normal. Effect size is the matched-pairs rank-biserial correlation. Each mean carries a 95 percent
confidence interval from a task-level bootstrap with 10000 resamples. A convergence claim is read from
both a not significant test and a confidence interval on the difference that is tight around zero, not
from the test alone. The paper reports several tests, so the headline effects are checked against a
Holm-Bonferroni correction: the exposure lifts and the retrieval gap, all at $p$ below 0.001, survive it.
The interaction at $p$ equal to 0.04 does not survive correction, so it is reported as suggestive, not
established.

Faithfulness depends on the judge and on the 0.5 entailment threshold, so both are probed. Recomputing
faithfulness at thresholds from 0.3 to 0.7 shifts the absolute values but not the conclusions. The
exposure lift holds at every threshold, for oracle and for naive, and the oracle to naive gap at 1500
characters never exceeds 0.016. Threshold choice therefore does not drive the result. To check that the
result is not an artifact of one judge, every cited claim is re-scored by a second independent judge,
HHEM-2.1-Open, a purpose-built grounding checker from a different model family that is used nowhere else in
the pipeline. Agreement with the primary judge and replication of the two findings under it are reported
in Section~\ref{sec:judgeval}.

\section{Expectations}
The design generates four predictions. E1: raising exposure raises faithfulness, though prior work is
split and a plateau or inversion is plausible for a small
model~\citep{webcites,contextutil}. E2: the exposure effect is larger with gold sources than with
retrieved sources, because noise in retrieved sources dilutes the signal~\citep{evidutil}. E3: retrieval,
not exposure, bounds coverage, because gold sits in the corpus yet retrieval misses much of it. E4:
repair after generation converges across variants and does not beat plain retrieval on trustworthy
coverage, because repair acts after the grounding is already lost~\citep{scorecorrect,selfcorrectillusion}.

\section{Results}
\begin{figure}[t]
\centering
\includegraphics[width=0.86\textwidth]{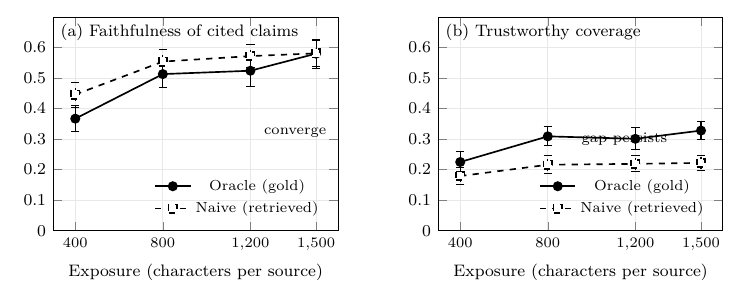}
\caption{Two levers on different outcomes, from the eight runs of Table~\ref{tab:config}. (a) Faithfulness
of cited claims rises with exposure, and the oracle and naive arms converge near 0.58, so faithfulness is
bound by exposure, not by whether the cited source is correct. (b) Trustworthy coverage stays split by
source quality and nearly flat in exposure, so coverage is bound by retrieval. Solid is oracle, dashed is
naive. Whiskers are bootstrap 95 percent confidence intervals over the 63 tasks.}
\label{fig:levers}
\end{figure}
\begin{table}[t]
\centering
\caption{Exposure by source quality at long output, 63 tasks each. Faith is cited claim faithfulness,
Rec is citation recall, TC is trustworthy coverage (Faith times Rec), Unc is the uncited claim rate,
Tok is mean output tokens. Precision is 1.00 for oracle by construction and 0.66 to 0.69 for naive.}
\label{tab:main}
\renewcommand{\arraystretch}{1.2}
\begin{tabular}{@{}llccccc@{}}
\toprule
Source & Exposure & Faith & Rec & TC & Unc & Tok \\
\midrule
Oracle & 400 ch  & 0.367 & 0.625 & 0.225 & 0.635 & 1380 \\
Oracle & 800 ch  & 0.513 & 0.627 & 0.309 & 0.579 & 1463 \\
Oracle & 1200 ch & 0.524 & 0.605 & 0.301 & 0.578 & 1541 \\
Oracle & 1500 ch & 0.580 & 0.604 & 0.328 & 0.578 & 1544 \\
Naive  & 400 ch  & 0.446 & 0.403 & 0.179 & 0.512 & 1439 \\
Naive  & 800 ch  & 0.554 & 0.392 & 0.216 & 0.529 & 1571 \\
Naive  & 1200 ch & 0.572 & 0.394 & 0.219 & 0.576 & 1623 \\
Naive  & 1500 ch & 0.581 & 0.396 & 0.222 & 0.538 & 1674 \\
\bottomrule
\end{tabular}
\end{table}

\begin{table}[t]
\centering
\caption{Paired significance of the main claims. Each row is a Wilcoxon signed-rank test across the 63
shared tasks. $\Delta$ is the mean paired difference, with a bootstrap 95 percent interval. $r$ is the
rank-biserial effect size.}
\label{tab:stats}
\renewcommand{\arraystretch}{1.25}
\footnotesize
\setlength{\tabcolsep}{3pt}
\begin{tabular}{@{}lcccc@{}}
\toprule
Comparison & $\Delta$ & 95\% CI & $p$ & $r$ \\
\midrule
Exposure on faith, oracle        & $+0.213$ & $[.17,.26]$ & $<.001$ & $0.88$ \\
Exposure on faith, naive         & $+0.135$ & $[.08,.19]$ & $<.001$ & $0.63$ \\
Faith, oracle vs naive at 1500   & $+0.001$ & $[-.05,.05]$ & $.86$ & $-0.03$ \\
Exposure on coverage, naive      & $+0.043$ & $[.02,.07]$ & $<.001$ & $0.58$ \\
Recall, oracle vs naive at 1500  & $+0.209$ & $[.16,.26]$ & $<.001$ & $0.98$ \\
Interaction, gain oracle vs naive & $+0.078$ & $[.01,.14]$ & $.04$ & --- \\
\bottomrule
\end{tabular}
\end{table}

\subsection{Faithfulness is a measure relative to the cited source}
Faithfulness scores each cited claim against the source it cites, not against the gold. This matters for
reading Table~\ref{tab:main}. At 400 characters the naive arm shows higher faithfulness, 0.446, than the
oracle arm, 0.367. This is not naive quality winning. The oracle arm is forced to cite the gold papers,
so it has high recall, 0.625, but is left making harder claims it cannot fully support at low exposure.
The naive arm cites what it retrieved and saw, so a larger share of its claims match the thing it cited,
even when that thing is not gold, which shows in its lower precision, 0.66, and low recall, 0.403. Oracle
precision is 1.00 by construction, since the oracle arm can only cite gold papers, so it is not a quality
signal and is not compared with the naive value. The honest number across arms is trustworthy coverage,
where the oracle arm already leads, 0.225 against 0.179, because its recall carries it. Faithfulness alone
can mislead, so it is always reported with recall and trustworthy coverage.

\subsection{Exposure lifts faithfulness, on gold and on retrieved sources}
Raising exposure lifts faithfulness in both arms, shown as a four-point curve in Fig.~\ref{fig:levers}(a).
On gold sources faithfulness rises 0.367, 0.513, 0.524, 0.580 across 400, 800, 1200, and 1500 characters;
on retrieved sources it rises 0.446, 0.554, 0.572, 0.581. Most of the gain comes in the first step, from
400 to 800 characters, 0.15 on gold and 0.11 on retrieved, and the curve is flatter after. The shape is not
perfectly smooth: on gold sources it plateaus from 800 to 1200 and rises again at 1500. Returns diminish
overall within the feasible range, which is the property the recipe
depends on, rather than assuming it. The end-to-end gains, 0.213 on gold and 0.135 on retrieved, are both
significant, in Table~\ref{tab:stats}. The oracle gain has $p$ below 0.001 and a large effect,
rank-biserial 0.88. The naive gain has $p$ below 0.001 and a moderate effect, 0.63. E1 held,
and in the rising direction, against the plateau or inversion that WebCiteS and the small model context
work would predict for this regime~\citep{webcites,contextutil}. E2 held on the size of the effect,
suggestively. The gain is larger on gold sources, 0.213 against 0.135, an interaction of 0.078 with $p$
equal to 0.04 (Table~\ref{tab:stats}). This does not survive the multiple-comparison correction, so it is
read as suggestive, consistent with retrieval noise diluting the signal but not established.
The lift also arrives with more claims written, not fewer. The mean claim count rises with exposure while
faithfulness rises, so the gain is not a matter of the model writing less.

\subsection{The arms converge, so faithfulness is bound by exposure}
At high exposure the two arms meet. Naive faithfulness, 0.581, equals oracle faithfulness, 0.580. The
difference is 0.001, not significant, with $p$ equal to 0.86, and its 95 percent interval runs from minus
0.05 to plus 0.05. The convergence is a tested null, tight around zero, not an eyeballed one. Given enough
of whatever it cites, the 4B supports its claims equally well whether the source is gold or merely
retrieved. Faithfulness of cited claims is therefore bound by exposure, not by whether the cited source is
the correct one. This is the central finding, and it qualifies the view that models below 7B cannot use
context~\citep{contextutil}. On the axis of exposure per source, the 4B does convert more input into more
faithful grounding.

\subsection{Trustworthy coverage is bound by retrieval}
Trustworthy coverage tells the opposite story, in Fig.~\ref{fig:levers}(b). It stays split by source
quality and nearly flat in exposure. On gold sources it moves 0.225 to 0.328. On retrieved sources it
moves only 0.179 to 0.222, and stays low because citation recall holds near 0.40 whatever the exposure.
Recall is statistically flat in exposure for the naive arm, with $p$ equal to 0.29, while the gap in
recall between the oracle and naive arms at high exposure is 0.209 and highly significant, with $p$ below
0.001 and effect 0.98 (Table~\ref{tab:stats}). E3 held. The reason is the retrieval stage. The gold papers
all sit in the corpus, so coverage is not the limit, yet retrieval surfaces less than half of them and the
agent cites most of what it surfaces. Recall is lost before the generator ever writes. Exposure cannot fix
which sources are cited, only how well the cited ones are supported.

\subsection{Repair after generation converges and does not lift coverage}
\label{sec:repair}
A short output panel tested whether repairing the brief after it is written helps, across four repair
actions. Table~\ref{tab:repair} shows the result. All three repairs in the loop lift faithfulness over the
naive draft by dropping or reattaching citations on weak claims, but they converge: gating, revising, and reattributing
land within a narrow band, so the repair action does not matter. And none beats plain retrieval on
trustworthy coverage, which sits near 0.19 for every arm, because each repair trades citation recall down
as it trades faithfulness up. E4 held. Repair acts after the grounding is lost. Exposure acts before it,
which is why exposure moves the number that repair cannot. These runs use short output and low exposure,
so they are not comparable to the long output grid above. They isolate the effect of the repair action,
not of exposure.

\begin{table}[t]
\centering
\caption{Short output repair panel, 63 tasks each. Repair lifts faithfulness but the action does not
matter, and no arm beats the naive draft on trustworthy coverage. Values characterize the convergence;
see the log for full provenance.}
\label{tab:repair}
\renewcommand{\arraystretch}{1.2}
\begin{tabular}{@{}lcccc@{}}
\toprule
Arm & Prec & Rec & Faith & TC \\
\midrule
Naive draft   & 0.698 & 0.385 & 0.437 & 0.168 \\
Gate          & 0.770 & 0.267 & 0.689 & 0.184 \\
Revise        & 0.754 & 0.299 & 0.659 & 0.197 \\
Reattribute   & 0.765 & 0.275 & 0.672 & 0.185 \\
Oracle        & 1.000 & 0.565 & 0.289 & 0.163 \\
\bottomrule
\end{tabular}
\end{table}

\subsection{Cost}
The exposure lift is nearly free. On retrieved sources, raising exposure adds about 235 output tokens,
from 1439 to 1674, a small fraction of the draft budget, for a 0.135 gain in faithfulness. Reading more
of each source costs more input tokens per task, but on the target hardware the batch stayed within the
memory budget and task times held near 45 seconds, so the lever is practical.

\subsection{The findings survive a second judge}
\label{sec:judgeval}
To rule out a single-judge artifact, every cited claim was re-scored by HHEM-2.1-Open, an independent
grounding checker from a different model family. Across the 9226 shared claims the two judges agree
moderately: Spearman correlation 0.65, agreement on the support decision 74.9 percent, Cohen kappa 0.50,
with near-identical overall support rates (0.51 for the primary judge, 0.52 for HHEM), so there is no
systematic bias. The exposure lift replicates under HHEM: faithfulness rises from 0.42 to 0.65 on gold
sources and from 0.44 to 0.61 on retrieved sources, the same direction and shape as the primary judge. The
convergence is weaker under HHEM. At 1500 characters the oracle and naive arms sit at 0.65 and 0.61, a gap
of 0.04, against the near-exact 0.001 gap the primary judge reports. So the exposure lift is judge-robust,
while the exact convergence of the two arms is tight under the primary judge and only approximate under the
second.

\section{Discussion}
\subsection{Two levers on different outcomes}
The agent loses citation quality in two places. It loses faithfulness at generation when it cannot see
enough of each source, and it loses coverage at retrieval when the right sources are never surfaced. The
two levers act on different outcomes. Exposure moves faithfulness and leaves coverage flat. Retrieval
moves coverage and does not decide faithfulness, since at high exposure the retrieved arm grounds as well
as the gold arm. The levers are not perfectly independent within faithfulness itself. Exposure helps more
when the sources are gold, a suggestive interaction of 0.078 (Table~\ref{tab:stats}), because clean
sources reward the extra reading more than noisy ones do. The recipe still reads off cleanly. Raise per
source exposure first, because it is cheap and it takes faithfulness from about 0.45 to about 0.58 on real
retrieval. Then spend effort on retrieval recall, because it is the only remaining lever for trustworthy
coverage.

\subsection{What to build}
The result argues against reaching first for a bigger model or for a repair loop. A bigger model does not
help coverage that is lost in retrieval, and the oracle arm shows the grounding gap is not a raw capacity
gap once the model can see the source. A repair loop converges and does not lift trustworthy coverage. The
move this study demonstrates is to widen exposure per source, which lifts faithfulness across the feasible
range on a fixed 4B laptop. The second move, improving retrieval recall with multiple queries or hybrid
expansion, is indicated by the diagnosis but not tested here. Whether a recall gain propagates to
trustworthy coverage is the clear next experiment, and it is the natural place to bring in larger models.

\subsection{Why report the triple}
Faithfulness is relative to the cited source, so a high value with low recall means the agent is
confidently citing the wrong things. That pattern is exactly the naive arm at high exposure: faithfulness
0.581 with recall 0.396. A single citation score would read that as success. The triple of faithfulness,
recall, and trustworthy coverage keeps it honest.

\subsection{The regime is improvable, not deployable}
The absolute levels are modest. Faithfulness tops out near 0.58, so about 42 percent of cited claims are
still unsupported, and trustworthy coverage reaches only 0.22 to 0.33. This 4B agent is not yet a reliable
citer. The contribution is not a deployable system, but a map of where the reliability must come from:
exposure for faithfulness, retrieval for coverage.

\section{Limitations}
The study fixes one 4B generator, one seed, and one corpus of research papers, so the numbers are a
characterization of this regime, not a universal curve. The confidence intervals cover task sampling, not
generator stochasticity, because the seed is fixed, so whether the small convergence gap holds across
seeds is untested. Gold is the set of papers a source paper actually cited, which is a proxy for what a
faithful answer must cite, not a hand-checked ground truth, and it inherits the citing habits of real
authors. Faithfulness is scored by an NLI judge against the cited source, which measures support, not
human judged correctness. The conclusions hold across entailment thresholds from 0.3 to 0.7 and replicate
under a second independent judge (Section~\ref{sec:judgeval}), but neither judge is calibrated against
human labels on this corpus, so absolute agreement with a human reader is unmeasured. The oracle arm supplies gold papers that are present in the corpus, an upper bound on source quality,
not a claim about perfect real retrieval. The exposure lever is a leading window, so the amount of source
shown is confounded with position. The first characters carry the title and abstract, and more characters
reach deeper, so part of the lever is whether the model reads past the abstract. A mid document or random
window control would separate amount from position. The exposure grid is run at long output only, so the
interaction with short output is not measured here. The retrieval lever is diagnosed, not moved: no recall
intervention is run, so the second half of the recipe is indicated rather than validated. A direct
comparison against other agent designs, at larger model sizes, is left to future work. This study
characterizes one fixed 4B, it does not rank systems.

\section{Conclusion}
For a 4B research agent on a laptop, citation quality splits into two levers that act on different
outcomes. Exposure bounds faithfulness: give the model enough of each source and it supports its claims as
well on retrieved sources as on gold ones, converging near 0.58 under the primary judge, with a second
judge confirming the lift and leaving a small residual gap. Retrieval bounds coverage: trustworthy
coverage stays near 0.22 on real retrieval at any exposure, because recall is set before generation. The
extra exposure costs about 235 tokens. The build order follows: exposure first, cheaply, then retrieval
recall. The absolute levels stay modest, so the contribution is a map of where reliability must come from,
not a deployable citer.

\section*{Availability}
\availabilitytext

\bibliographystyle{tmlr}
\bibliography{refs}

\appendix
\begin{center}\large\bfseries\sffamily Appendix\end{center}
\section{Full per-cell results}
Table~\ref{tab:appfull} gives every metric for all eight cells of the exposure by source-quality grid,
63 tasks each. Faith is cited-claim faithfulness, Prec and Rec are citation precision and recall, TC is
trustworthy coverage, Unc is the uncited-claim rate, Claims is the mean cited claim count, Tok is mean
output tokens. Precision is 1.00 for oracle by construction.

\begin{table}[h]
\centering
\caption{Full per-cell results. Long output, 63 tasks each.}
\label{tab:appfull}
\small
\setlength{\tabcolsep}{4pt}
\begin{tabular}{@{}llccccccc@{}}
\toprule
Source & Exp & Faith & Prec & Rec & TC & Unc & Claims & Tok \\
\midrule
Oracle & 400  & 0.367 & 1.00 & 0.625 & 0.225 & 0.635 & 14.7 & 1380 \\
Oracle & 800  & 0.513 & 1.00 & 0.627 & 0.309 & 0.579 & 17.2 & 1463 \\
Oracle & 1200 & 0.524 & 1.00 & 0.605 & 0.301 & 0.578 & 18.1 & 1541 \\
Oracle & 1500 & 0.580 & 1.00 & 0.604 & 0.328 & 0.578 & 18.0 & 1544 \\
Naive  & 400  & 0.446 & 0.660 & 0.403 & 0.179 & 0.512 & 19.2 & 1439 \\
Naive  & 800  & 0.554 & 0.669 & 0.392 & 0.216 & 0.529 & 19.9 & 1571 \\
Naive  & 1200 & 0.572 & 0.684 & 0.394 & 0.219 & 0.576 & 18.6 & 1623 \\
Naive  & 1500 & 0.581 & 0.687 & 0.396 & 0.222 & 0.538 & 20.7 & 1674 \\
\bottomrule
\end{tabular}
\end{table}

\section{Second judge: HHEM results and agreement}
Every cited claim was re-scored by HHEM-2.1-Open. Over the 9226 shared claims the two judges give Spearman
correlation 0.65, agree on the support decision 74.9 percent of the time, and reach a Cohen kappa of 0.50,
with near-identical overall support rates, 0.512 for the primary judge and 0.521 for HHEM. Table~\ref{tab:apphhem}
gives per-cell faithfulness under both judges. The exposure lift is present under both. The oracle to naive
gap at 1500 characters is 0.001 under the primary judge and 0.038 under HHEM.

\begin{table}[h]
\centering
\caption{Per-cell faithfulness under the primary judge (DeBERTa-NLI) and the second judge (HHEM).}
\label{tab:apphhem}
\small
\begin{tabular}{@{}llcc@{}}
\toprule
Source & Exposure & DeBERTa & HHEM \\
\midrule
Oracle & 400  & 0.367 & 0.419 \\
Oracle & 800  & 0.513 & 0.504 \\
Oracle & 1200 & 0.524 & 0.585 \\
Oracle & 1500 & 0.580 & 0.645 \\
Naive  & 400  & 0.446 & 0.438 \\
Naive  & 800  & 0.554 & 0.551 \\
Naive  & 1200 & 0.572 & 0.581 \\
Naive  & 1500 & 0.581 & 0.607 \\
\bottomrule
\end{tabular}
\end{table}

\section{Judge threshold sensitivity}
Faithfulness is the share of cited claims whose entailment probability against the cited source is at least
0.5. Table~\ref{tab:appthr} recomputes per-cell faithfulness at thresholds from 0.3 to 0.7. Absolute values
shift, but the exposure lift holds at every threshold in both arms, and the oracle to naive gap at 1500
characters stays small.

\begin{table}[h]
\centering
\caption{Per-cell faithfulness at entailment thresholds 0.3 to 0.7 (primary judge).}
\label{tab:appthr}
\small
\begin{tabular}{@{}llccccc@{}}
\toprule
Source & Exposure & 0.3 & 0.4 & 0.5 & 0.6 & 0.7 \\
\midrule
Oracle & 400  & 0.493 & 0.417 & 0.367 & 0.324 & 0.282 \\
Oracle & 800  & 0.607 & 0.565 & 0.514 & 0.466 & 0.416 \\
Oracle & 1200 & 0.643 & 0.583 & 0.525 & 0.483 & 0.446 \\
Oracle & 1500 & 0.687 & 0.631 & 0.582 & 0.549 & 0.507 \\
Naive  & 400  & 0.582 & 0.501 & 0.446 & 0.395 & 0.348 \\
Naive  & 800  & 0.641 & 0.591 & 0.554 & 0.517 & 0.465 \\
Naive  & 1200 & 0.675 & 0.620 & 0.572 & 0.537 & 0.492 \\
Naive  & 1500 & 0.674 & 0.622 & 0.581 & 0.533 & 0.494 \\
\bottomrule
\end{tabular}
\end{table}

\section{Recall decomposition}
Citation recall factors as retrieval recall times utilization, where retrieval recall is the share of gold
that retrieval surfaced and utilization is the share of surfaced gold that the agent cited. For the naive
arm, retrieval recall is 0.432 at every exposure, and utilization is between 0.91 and 0.93. So citation
recall near 0.40 is set by retrieval, not by the generator, and exposure does not move it. The gold papers
all sit in the corpus, so corpus coverage is complete and the loss is first-stage retrieval.

\section{Post-generation repair panel}
A separate short-output panel tested whether repairing the brief after it is written helps.
Table~\ref{tab:apprepair} gives the arms. All three in-loop repairs lift faithfulness over the naive draft
by dropping or re-citing weak claims, but they converge and none beats the naive draft on trustworthy
coverage. These runs use short output and low exposure and are not comparable to the long-output grid.

\begin{table}[h]
\centering
\caption{Short-output repair panel, 63 tasks each.}
\label{tab:apprepair}
\small
\begin{tabular}{@{}lcccc@{}}
\toprule
Arm & Prec & Rec & Faith & TC \\
\midrule
Naive draft & 0.698 & 0.385 & 0.437 & 0.168 \\
Gate        & 0.770 & 0.267 & 0.689 & 0.184 \\
Revise      & 0.754 & 0.299 & 0.659 & 0.197 \\
Reattribute & 0.765 & 0.275 & 0.672 & 0.185 \\
Oracle      & 1.000 & 0.565 & 0.289 & 0.163 \\
\bottomrule
\end{tabular}
\end{table}

\section{Corpus and task construction}
The substrate is built from the public DeepScholar-Bench dataset. Each task is one source paper, whose
title and abstract are the research question, and whose own reference list is the gold evidence set, an
average of 17.7 papers per task, ranging from 2 to 35. The corpus is the union of the 63 source papers and
every paper they cite, 1062 documents in all, frozen as one content-hashed snapshot (hash
\texttt{e77dc6fb8aee51a3}). Of these, 656 carry full body text fetched from arXiv and the rest carry a
title, an abstract, and a snippet. Every gold citation resolves to a corpus document by arXiv identifier or
by normalized title, so the gold set is a subset of the corpus by construction. The article text is not
redistributed. The released repository ships the paper list and a rebuild script that reconstructs the
corpus from DeepScholar-Bench and arXiv.

\section{Reproducibility and environment}
All runs use one generator, Qwen3-4B-Instruct-2507 at 8 bit, served by Ollama on an Apple M4 Pro with 24 GB
of memory, at temperature 0.2, a context window of 8192, and a fixed seed of 42. The retriever is hybrid,
BM25 with a nomic dense encoder, fused by reciprocal rank fusion. The primary judge is a DeBERTa-v3 NLI
cross encoder with a SummaC windowed maximum; the second judge is HHEM-2.1-Open. The exposure grid runs the
draft step at 400, 800, 1200, and 1500 characters per source, for the oracle and naive arms, at a 2000
token draft budget. Every run stores raw per-claim grades with judge identity and provenance, and every
table above regenerates from those stored records without re-running the agent.

\end{document}